\documentclass[journal]{IEEEtran}
\usepackage{amsmath}
\usepackage{graphicx}
\graphicspath{{imgs/}}
\usepackage{float}
\usepackage{multirow}
\usepackage{xcolor}
\usepackage{flushend}

\ifCLASSINFOpdf
\else
\fi
\begin{document}
%
\title{Multimodal AI on Wound Images and Clinical Notes for Home Patient Referral}
%
%
%

\author{Reza~Saadati Fard,
        Emmanuel ~Agu,
        Palawat ~ Busaranuvong,
        Deepak ~ Kumar,
        Shefalika~Gautam,
        Bengisu~Tulu, 
        and~Diane~Strong
        

\thanks{R. Saadati Fard, E. Agu, and D. Kumar are with the Department of Computer Science at Worcester Polytechnic Institute, Worcester, MA 01609, USA (email: rsaadatifard@wpi.edu)}
\thanks{B. Busaranuvong, and S. Gautam are with the Department of Data Science at Worcester Polytechnic Institute, Worcester, MA 01609, USA}
\thanks{B. Tulu, and D. Strong are with the Business School at Worcester Polytechnic Institute, Worcester, MA 01609, USA}
}

\maketitle

\begin{abstract}
Chronic wounds affect 8.5 million Americans, especially the elderly and patients with diabetes. Since such wounds can take up to nine months to heal, regular care is crucial to ensure proper healing, and prevent severe outcomes such as limb amputations. However, many patients receive care in their homes from visiting nurses and caregivers with variable wound expertise, resulting in inconsistent care. Problematic, non-healing wounds should be referred to experts in wound clinics to avoid adverse outcomes such as limb amputations. Unfortunately, due to the lack of wound expertise, referral decisions made in non-clinical settings can be erroneous, delayed or unnecessary. \\ 
\indent This paper proposes the Deep Multimodal Wound Assessment Tool (DM-WAT), a novel machine learning framework to support visiting nurses in deciding whether to refer chronic wound patients to see a wound specialist. DM-WAT analyzes smartphone-captured wound images and clinical notes from Electronic Health Records (EHRs) to recommend whether a patient should be referred to a wound specialist. DM-WAT extracts visual features from wound images using  DeiT-Base-Distilled, a Vision Transformer (ViT) architecture. Distillation-based training facilitates representation learning and knowledge transfer from a larger teacher model to DeiT-Base so that DeiT-Base-Distilled performs well on our small wound image dataset of 205 wound images. DM-WAT extracts text features from clinical notes using DeBERTa-base, which improves context comprehension by disentangling content and position information from clinical notes. DeBERTa-base’s disentangled attention mechanism ensures robust extraction of complex syntactic and semantic dependencies from clinical text. Visual and text features are combined using an intermediate fusion approach. To overcome the challenges posed by a small and imbalanced dataset, DM-WAT integrates image and text augmentation alongside transfer learning via pre-trained feature extractors to achieve high performance. In rigorous evaluation, DM-WAT achieved an accuracy of 77\% ± 3\% and an F1 score of 70\% ± 2\%, outperforming the prior state of the art and all baseline single-modality and multimodal approaches. Additionally, to enhance the interpretability and trust in DM-WAT's recommendations, the Score-CAM and Captum interpretation algorithms provided insights into the specific parts of the image and text inputs that the model focused on during decision-making.

\end{abstract}


%


\section{Introduction}
%
%
%
%

\IEEEPARstart{C}{hronic} wounds affect approximately 8.2 million people in the United States~\cite{jarbrink2016prevalence, sen2021human}, and have prolonged healing times and significant economic costs \cite{1frykbergrobert2015challenges, 2sen2019human}. These wounds are particularly common among the elderly and patients with conditions such as diabetes, contributing to an annual healthcare burden ranging from \$25 \text{to} \$96.8 \text{billion} \cite{jarbrink2016prevalence, sen2021human}. Timely and accurate wound assessment is crucial, as it determines whether a patient’s current treatment plan is effective or needs modification \cite{4nguyen2020machine}. However, the shortage of wound specialists has resulted in a situation where much of the follow-up care is provided by visiting nurses with no specialized wound training in home settings, leading to inconsistent and non-standardized wound management. Specifically, patients with problematic wounds who are receiving home care should be referred to experts in a clinic to avoid adverse outcomes such as limb amputations. However, these referral decisions are sometimes delayed or inaccurate, which in turn can lead to avoidable limb amputations, and even fatalities \cite{5jarbrink2017humanistic}. Given these challenges, there is a clear need for solutions that can support visiting nurses with no specialized wound training in making informed and timely wound care decisions.

Previous work in wound assessment has included traditional clinical tools such as the Photographic Wound Assessment Tool (PWAT), which provides structured rubrics to evaluate wound healing progress \cite{WAT2hon2010prospective, PWATthompson2013reliability}. However, these tools are often manual, time-intensive, and rely on the expertise of trained specialists, which may not always be available to visiting nurses. In recent years, AI-driven solutions have been introduced to address these limitations. Some approaches have utilized image-based models, primarily convolutional neural networks (CNNs), to automate wound classification based on visual data\cite{cnn1zhang2022survey, cnn2rostami2021multiclass}. CNNs have demonstrated success in identifying wound types and stages, including diabetic ulcers and pressure ulcers, providing reliable tools for clinical applications \cite{cnn2rostami2021multiclass}.  More recently, Vision Transformers (ViTs) have emerged as a powerful alternative to CNNs for medical image analysis \cite{vitdosovitskiy2020image, vitApp1mohan2024vision}. Unlike CNNs, ViTs use self-attention mechanisms to capture global dependencies across an image, enabling them to excel in tasks requiring detailed spatial analyses~\cite{vitdosovitskiy2020image, vitApp1mohan2024vision}. This advancement suggests that ViTs may be effective for wound image assessment. While image-based models are effective, they often lack the ability to incorporate contextual information from clinical notes, which is crucial for a comprehensive understanding of wound conditions. Other studies have explored text-based models such as Term Frequency-Inverse Document Frequency (TF-IDF) \cite{tfidf1lasko2013computational} and Hierarchical Attention Networks (HAN) \cite{4nguyen2020machine}. Moreover, language models such as Bidirectional Encoder Representations from Transformers (BERT)~\cite{bertdevlin2018bert} have proven effective in processing and classifying Electronic Health Record (EHR) data. Despite the effectiveness of text-based algorithms, these models are limited in their ability to analyze visual data, which is often critical for chronic wound evaluations. Although multimodal AI methods that combine visual and textual data have shown promise in various healthcare applications \cite{yildirim2024multimodal, hartsock2024vision, 16mombini2020design}, their use in wound assessment have not been explored extensively.

\begin{figure*}[h]
\centering
\includegraphics[width=1 \textwidth]{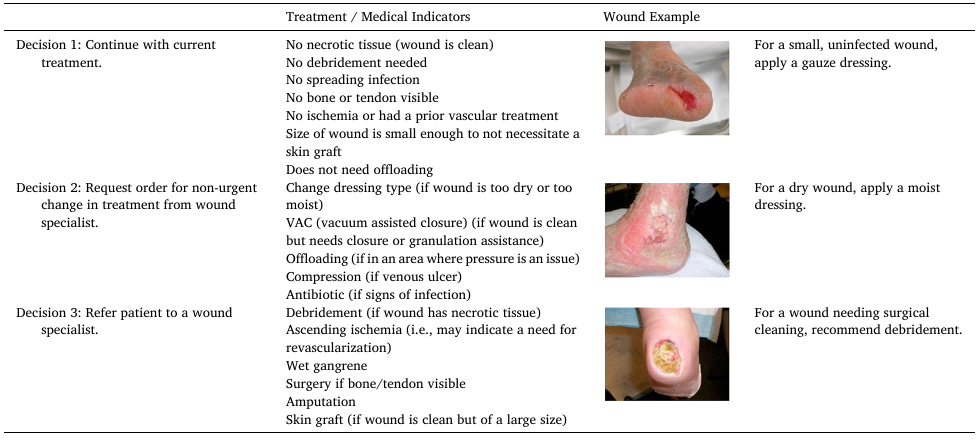}
\caption{Example of Wound Image from Dataset. Each image corresponds to one of the three referral decision categories along with descriptive clinical notes \cite{4nguyen2020machine}.}
\label{fig:example_image}
\end{figure*}

In this work, we propose the \textbf{Deep Multimodal Wound Assessment Tool (DM-WAT)}, an AI framework that utilizes both wound images and clinical notes to provide accurate referral recommendations. Our data consists of 205 wound images and associated clinical notes collected from UMass Memorial Medical Center, each categorized by wound care specialists into one of three referral classes: (1) \textit{Continue Treatment}, (2) \textit{Change Treatment Non-Urgently}, and (3) \textit{Change Treatment Urgently} (See figure \ref{fig:example_image}). This multimodal dataset allows DM-WAT to address the limitations of image-only or text-only approaches by leveraging both visual and textual information.

The proposed DM-WAT framework is designed to process and analyze both image and text data by combining advanced deep learning models for each modality and fusing their outputs. For wound images, we use the DeiT-Base-Distilled model, a Vision Transformer (ViT) architecture known for its efficient use of data and ability to capture intricate visual details \cite{deittouvron2021training}. The DeiT-Base-Distilled model is pre-trained on a large dataset of general images, allowing it to extract relevant visual features from wound images, even with our relatively small dataset. For textual data, we employ the DeBERTa-base model, a BERT-based language model that provides high-quality contextual embeddings from clinical notes \cite{debertahe2020deberta}. DeBERTa is effective at capturing subtle textual information that reflects wound severity and treatment recommendations. After extracting features from each modality intermediate fusion is used to concatenate the  visual and textual features into a combined representation. This fused vector is then used by the final classifier to predict the referral decision based on a comprehensive understanding of both the wound’s appearance and clinical context.

Our contributions are as follows:
\begin{itemize}
    \item \textbf{Automatic feature extraction using pre-trained Networks}: DeiT-Base-Distilled captures complex visual features from images, while DeBERTa-base generates rich contextual embeddings from clinical notes, improving the model’s ability to understand both visual and textual cues for accurate classification.
    \item \textbf{Synthetic data generation to address missing or sparse data}: Synthetic data were generated and utilized to augment both data modalities, enhancing model robustness and generalizability. Specifically, classic techniques such as rotation, flipping, and brightness adjustments were used for wound image augmentation, while GPT-4 was employed to generate synthetic clinical notes.
    \item \textbf{Intermediate fusion to combine visual and textual features}: Intermediate fusion was applied to combine visual and textual features, significantly enhancing prediction accuracy for all three target referral classes. A Support Vector Machine (SVM) classifier was utilized to classify the fused representation of features extracted from DeiT-Base-Distilled and DeBERTa-base into the three target referral classes.
    \item \textbf{Interpretability for Clinical Application}: Interpretability methods such as Score-CAM and Captum were employed to provide insights into the specific parts of image and text inputs that influenced model decisions. These methods enhanced transparency and trust, thereby increasing clinician confidence in DM-WAT's recommendations.
    \item \textbf{Rigorous evaluation}: DM-WAT achieved an accuracy of 77\% ± 3\% and an F1 score of 70\% ± 2\%, outperforming the prior state of the art and all baseline single-modality and multimodal approaches.
\end{itemize}

The rest of this paper is structured as follows: Section \ref{sec:related_work}  reviews related work, Section \ref{sec:methodology} describes our methodology including the dataset and DM-WAT architecture, Section \ref{sec:evaluation} presents evaluation experiments and results, Section \ref{sec:discussion} discusses findings, and Section \ref{sec:conclusion} concludes with future directions.

\section{Related work} \label{sec:related_work}

The integration of machine learning (ML) into healthcare has gained significant traction, particularly for wound care and assessment. Paper-based wound assessment rubrics, such as the Leg Ulcer Measurement Tool \cite{WAT1woodbury2004development}, Pressure Ulcer Scale for Healing \cite{WAT2hon2010prospective}, and the Photographic Wound Assessment Tool (PWAT) \cite{PWATthompson2013reliability}, have provided guidance for non-specialist clinicians. These tools enable structured evaluation and scoring of wound characteristics, including size, depth, and tissue type, thus supporting effective monitoring and treatment over time~\cite{4nguyen2020machine, WAT1woodbury2004development}. However, such paper-based rubrics still rely on human observation, making them prone to inconsistency and subjective interpretation, and can be time-consuming to utilize. Additionally, as these manual wound assessment rubrics are not data-driven and do not learn from prior wound assessment data, they are unable to leverage the large amount of classified data already assessed by experts, which could otherwise improve wound assessment through machine learning integration \cite{PWATthompson2013reliability}.

Machine learning algorithms have emerged as promising solutions to address these limitations by enabling automated, consistent wound assessments and decision support. ML algorithms for wound assessment can be broadly categorized into three types: image-based, text-based, and multimodal approaches. Each category offers unique advantages and applications for wound care.

\subsection{Image-based machine learning for wound assessment}

Image-based ML models such as Convolutional Neural Networks (CNNs), have shown high accuracy in classifying wound types and stages. For instance, Rostami et al. \cite{cnn2rostami2021multiclass} developed an ensemble CNN model to classify wound images into categories such as surgical, diabetic, and venous ulcers, achieving up to 96.4\% accuracy in binary classification. Vision Transformer (ViT) models, a more recent development, have also demonstrated promising results in image classification tasks in wound care. Mohan et al. \cite{vitApp1mohan2024vision} used a ViT model for diabetic foot ulcer detection, achieving 98.58\% accuracy and outperforming traditional CNN-based methods \cite{vitdosovitskiy2020image}. These examples illustrate the potential of image-based ML models to provide robust, automated wound assessments that are faster and typically more accurate than manual assessment by humans even when guided by paper-based rubrics.

\subsection{Text-based machine learning for wound assessment}
Text-based models leverage Electronic Health Records (EHR) and clinical notes to capture detailed patient history, enabling a comprehensive wound analysis. Advanced Natural Language Processing (NLP) models, such as Bidirectional Encoder Representations from Transformers (BERT) and its medical-specific variant ClinicalBERT, have shown success in analyzing EHR data by capturing nuanced contextual information in clinical notes \cite{10huang2019clinicalbert}. For example, ClinicalBERT, fine-tuned on EHRs, has been applied to hospital readmission prediction with superior performance over traditional NLP models~\cite{bertdevlin2018bert}. This capability enables accurate identification of key factors in wound progression and treatment response, making text-based ML an essential component for understanding wound conditions from a clinical perspective.

\subsection{Multimodal machine learning for wound referral recommendation}
Recent studies have shown that multimodal approaches, which integrate both image and text data, provide a holistic view of the wound condition and enhance decision-making accuracy \cite{16mombini2020design, yildirim2024multimodal}. Nguyen et al. \cite{4nguyen2020machine} proposed a multimodal model combining image and textual features for wound care decision-making, achieving improved accuracy in referral recommendations by leveraging PWAT scores along with clinical notes. Holly et al. \cite{4nguyen2020machine} applied a multimodal model using a Hierarchical Attention Network (HAN) to process clinical notes and visual features, enhancing the reliability of referral decisions for chronic wounds. These findings highlight the potential of multimodal approaches to utilize both visual and textual cues, leading to more accurate wound assessments.

Our work builds on these foundations by introducing the Deep Multimodal Wound Assessment Tool (DM-WAT), which combines image features extracted with the DeiT-Base-Distilled model and text features derived from DeBERTa. This model is designed to enhance decision support for wound care, particularly by improving the accuracy of referral recommendations for chronic wounds. Leveraging these state-of-the-art architectures, our approach provides reliable, actionable insights that assist clinicians in making timely and effective wound care decisions.

\section{Methodology} \label{sec:methodology}

This methodology section details the  Deep Multimodal Wound Assessment Tool (DM-WAT) architecture, dataset characteristics and limitations, data augmentation strategies, feature extraction methods, and classification techniques, along with interpretation algorithms that enhance the transparency of model predictions.

\subsection{Dataset}

The dataset used in this study consists of 205 wound images obtained from patients at UMass Memorial Medical Center \cite{4nguyen2020machine, 16mombini2020design}. These images were carefully selected to represent various wound types and severity levels, ensuring a comprehensive depiction of the wound spectrum. Each image was reviewed and labeled by two wound specialists—Expert 1, a plastic surgeon, and Expert 2, a dually credentialed podiatric surgeon and vascular nurse practitioner. These experts both provided ground truth labels by categorizing each wound into one of three treatment decision categories based on their clinical judgment and expertise:

\begin{enumerate}
    \item \textbf{Continue Current Treatment}: Indicating that the current treatment plan for the patient's wound was deemed appropriate and effective by the experts, with no changes needed.
    \item \textbf{Change Treatment Non-Urgently}: Suggesting that modifications to the treatment plan were necessary, although urgent intervention was not required. This category includes cases where adjustments to wound care protocols or additional medical interventions might support optimal healing.
    \item \textbf{Change Treatment Urgently}: Referring to wounds needing immediate intervention due to factors such as infection, worsening condition, or lack of response to current treatment. Urgent changes aim to prevent further complications and adverse outcomes, such as limb amputation, and to promote healing.
\end{enumerate}

Alongside each image, the specialists provided detailed textual descriptions, including insights into wound characteristics and rationale behind their referral decisions. Figure \ref{fig:example_image} shows examples of the three referral decision categories, illustrating the diversity of wound types and the types of clinical notes used to describe them.

\subsection{Challenges and strategies to mitigate dataset issues} \label{addressDataProblem}

\begin{figure}[htbp]
\centering
\includegraphics[width=0.5\textwidth]{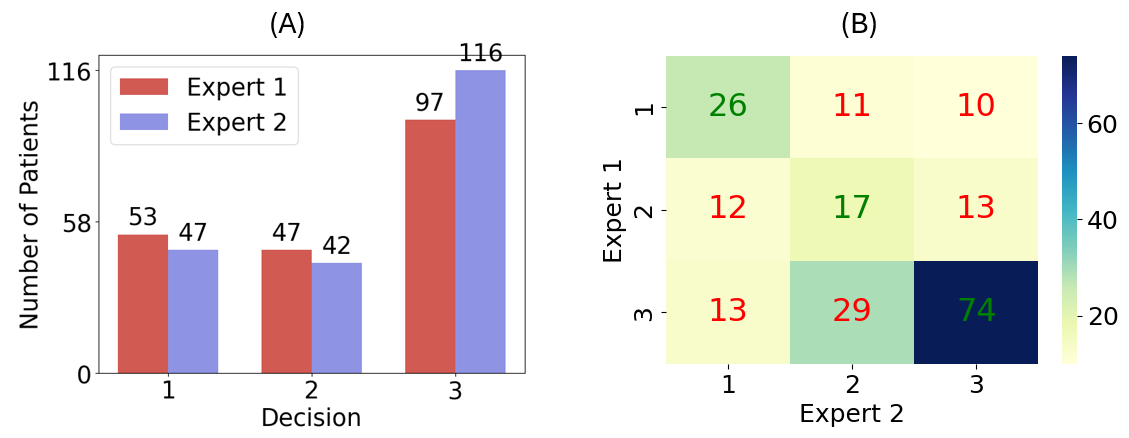}
\caption{\textbf{Expert Decision Analysis} \textbf{(A)} The bar chart illustrates the imbalance in referral decisions, with most cases falling under urgent referral. \textbf{(B)} The confusion matrix displays agreement (diagonal) and disagreement (off-diagonal) between the experts' decisions, highlighting inconsistencies in the labeling.}
\label{fig:data_limitation}
\end{figure}

The dataset presents several challenges, including its small size, class imbalance, and occasionally contradictory labels from the experts. Addressing these limitations is critical for training reliable and accurate machine learning models. The following strategies were implemented to mitigate these issues:

\begin{itemize}
    \item \textbf{Small Dataset}: With only 205 images, the dataset’s size is a major limitation for deep neural network (DNN) performance. To address this, data augmentation techniques were employed to increase the size and variability of the dataset, facilitating robust model training on limited data. Traditional augmentation methods, such as rotations, flips, and random cropping, were applied to wound images \cite{perez2018data}. For textual data, GPT-4 was used to generate additional wound descriptions based on the clinical context, enhancing the dataset with synthetic but realistic text \cite{openai2023gpt4}.

    \item \textbf{Imbalanced Dataset}: The majority of cases in our dataset belonged to the urgent treatment class. This imbalance was because the dataset was sourced from a wound clinic, which generally treated more severe wound cases. Figure \ref{fig:data_limitation}(A) shows a bar chart depicting this imbalance across the three categories. To mitigate data imbalance, data augmentation was used to upsample underrepresented categories, aiming to create a more balanced dataset for model training.

    \item \textbf{Contradictory Labels}: There were instances where the two experts provided conflicting recommendations for the same case, introducing inconsistencies into the dataset. To prioritize patient safety, a conservative final decision, $\text{dec}_{\text{final}}$, was adopted by selecting the higher (more urgent) of the two expert recommendations, as defined in Equation~\eqref{eq:final_decision}:
    
    \begin{equation} \label{eq:final_decision}
    \text{dec}_{\text{final}} = \max(\text{dec}_{\text{exp1}}, \text{dec}_{\text{exp2}}),
    \end{equation}
    
    where $\text{dec}_{\text{exp1}}$ and $\text{dec}_{\text{exp2}}$ represent the decisions made by Expert 1 and Expert 2, respectively. Choosing the more urgent recommendation ensures that the model errs on the side of caution. Figure~\ref{fig:data_limitation}(B) illustrates the agreement (diagonal) and disagreement (off-diagonal) between the experts' decisions, providing insight into the consistency of their labeling.

\end{itemize}

\subsection{DM-WAT architecture}

\begin{figure*}[t]
\centering
\includegraphics[width=0.75\textwidth]{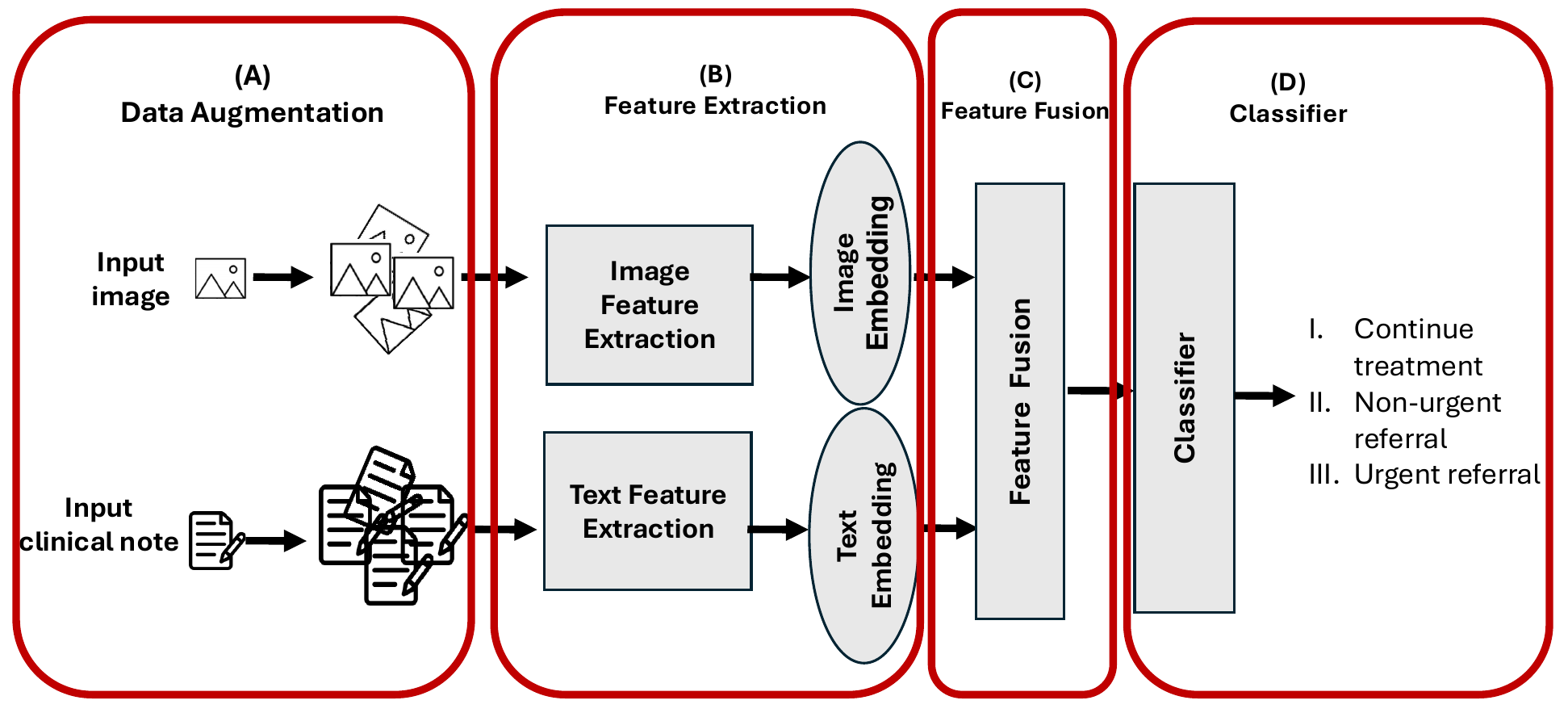}
\caption{\textbf{DM-WAT framework}: (A) Data augmentation, (B) feature extraction using deep neural networks, (C) intermediate fusion of features, and (D) classification into three referral categories: continue treatment, non-urgent referral, or urgent referral.}
\label{fig:pipeline}
\end{figure*}

Figure \ref{fig:pipeline} presents an overview of the DM-WAT framework, which consists of four main stages: data augmentation, feature extraction, multimodal fusion, and classification.

\subsubsection{Image augmentation}

\begin{figure}
\centering
\includegraphics[width=.5\textwidth]{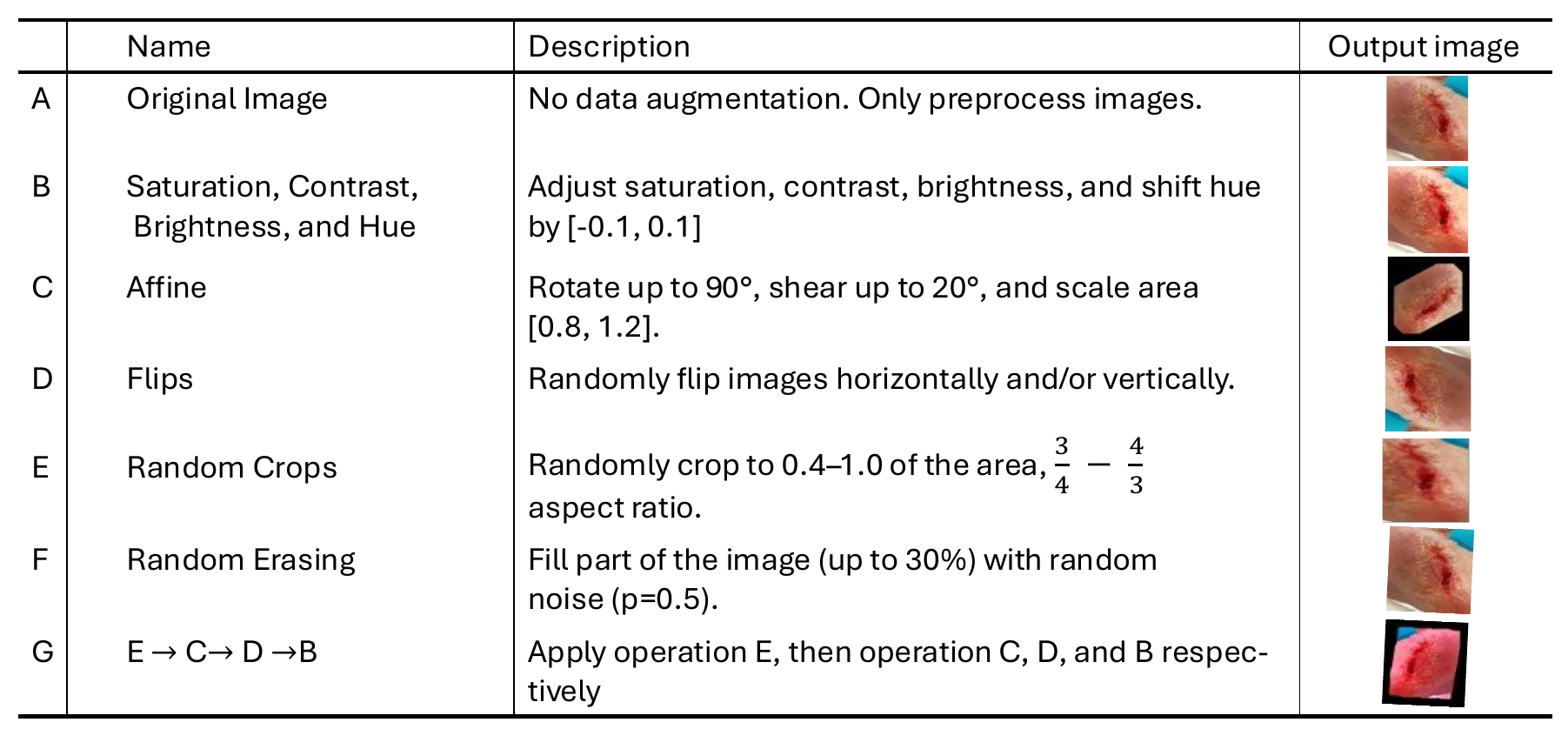}
\caption{Image augmentation operations with  visual examples}
\label{fig:augmentation_scenarios_pdf}
\end{figure}

To increase the diversity of the wound image dataset, classic augmentation techniques, such as rotations, flips, and random crops, were applied to each wound image \cite{20thompson2013reliability}. These augmentations help increase the size of the data, address class imbalance and add variability to the limited dataset. Figure \ref{fig:augmentation_scenarios_pdf} provides examples of these transformations.

\subsubsection{Text augmentation} \label{secTxtAug}

\begin{figure}
\centering
\includegraphics[width=\columnwidth]{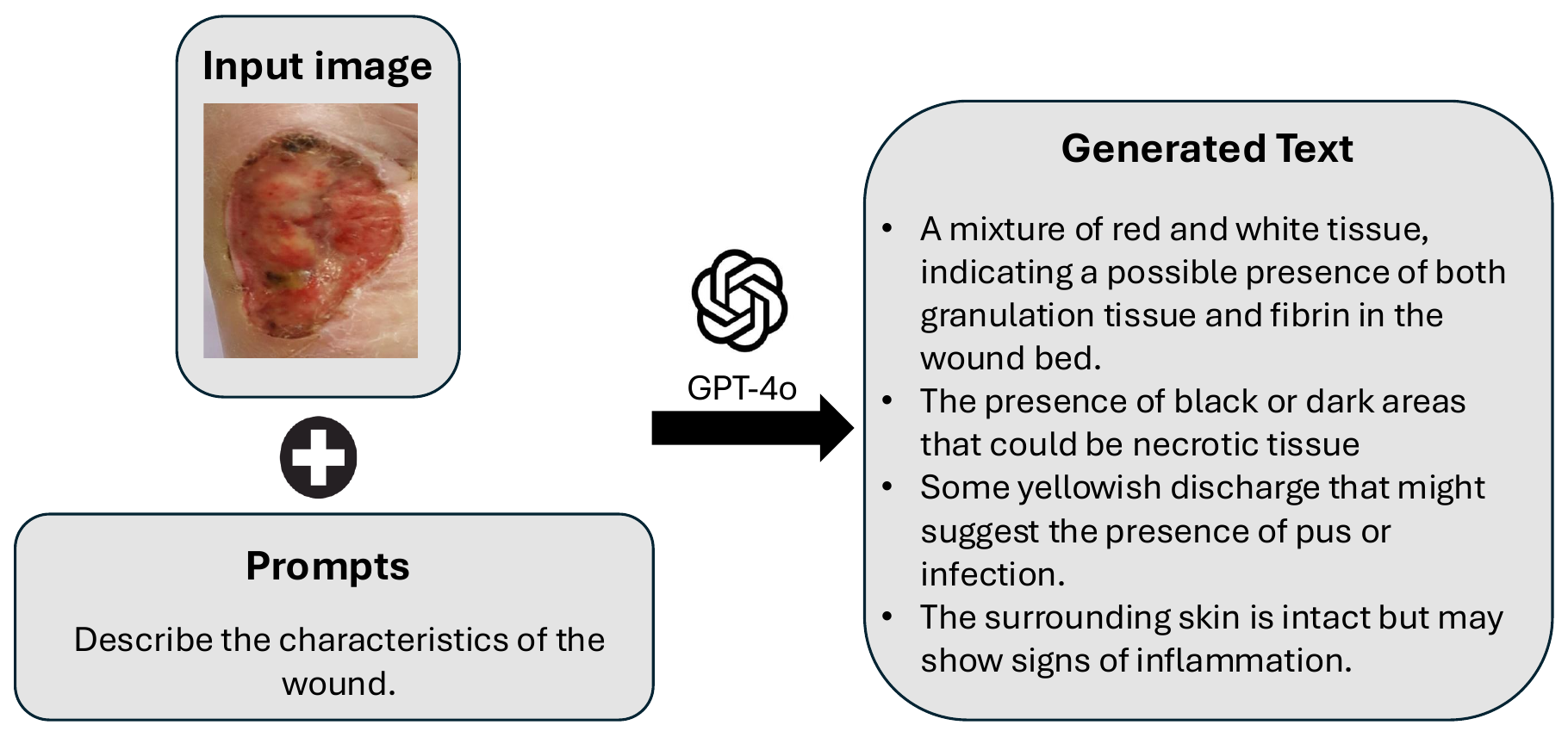}
\caption{Example of GPT-4 generated text for Wound Descriptions}
\label{fig:aug_text}
\end{figure}

GPT-4 was employed to augment textual data by generating synthetic descriptions based on visual prompts, thereby enriching the dataset with additional clinical descriptions. This synthetic text was combined with expert-provided notes to enable the model to learn from a more diverse set of wound descriptions. Figure \ref{fig:aug_text} illustrates an example of GPT-4-generated text.

\subsection{Feature extraction}

\subsubsection{Visual feature extraction utilizing DeiT-Base-Distilled}

\begin{figure}
\centering
\includegraphics[width=0.4\textwidth]{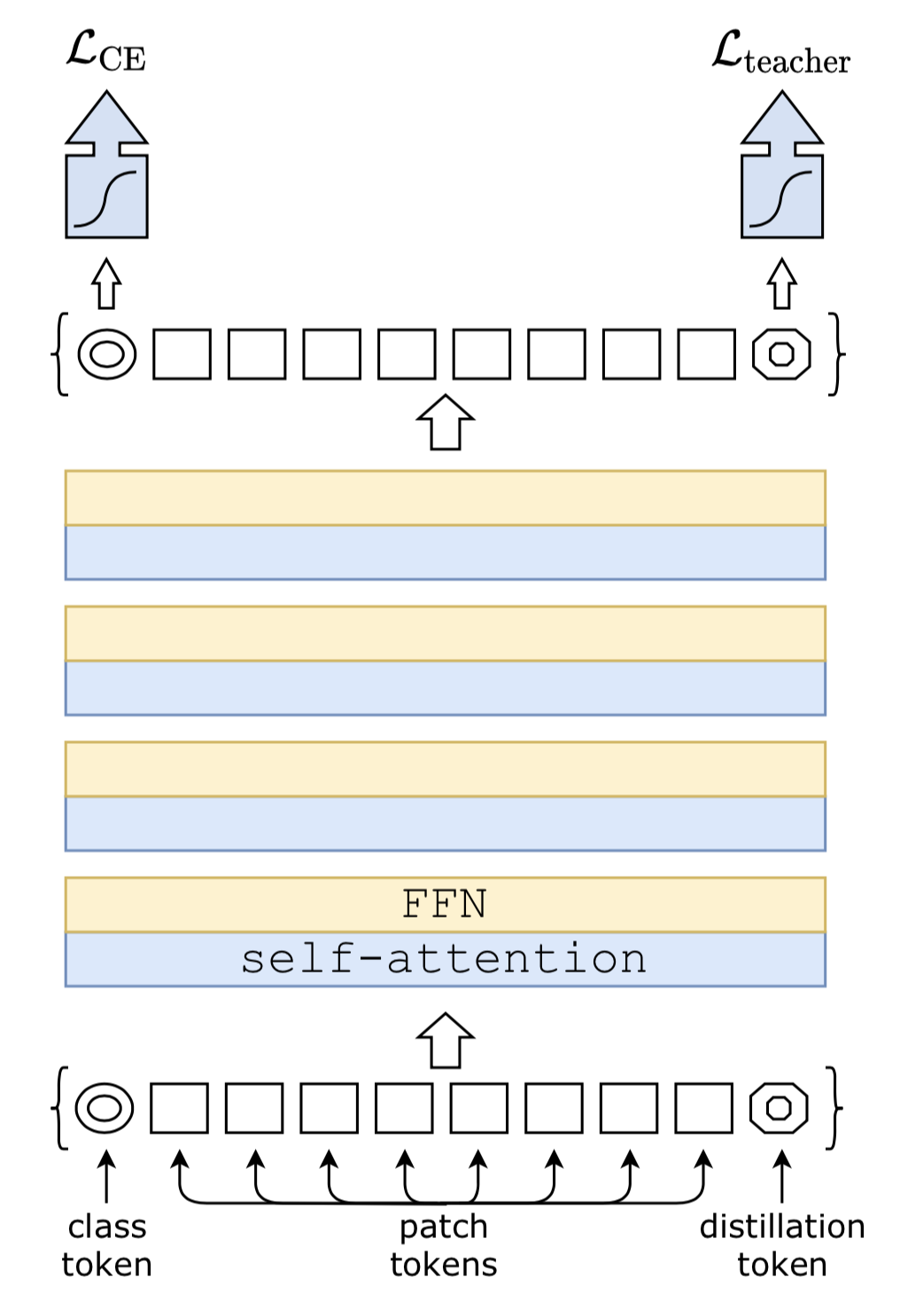}
\caption{DeiT-Base-Distilled Architecture with Class and Distillation Tokens \cite{deit_keras}.}
\label{fig:deit_architecture}
\end{figure}

DeiT-Base-Distilled is a data-efficient vision transformer architecture designed to extract detailed visual features, even from small datasets. DeiT leverages knowledge distillation from a teacher network, which enhances its generalization capability \cite{deittouvron2021training}. The model processes wound images by embedding image patches into a sequence of tokens. It uses both class and distillation tokens to improve feature extraction, as shown in Figure \ref{fig:deit_architecture}.

\textbf{Knowledge distillation} 

Knowledge distillation enables DeiT-Base-Distilled to learn from both the actual labels and the teacher model’s predictions, enhancing its ability to generalize effectively. The total loss for training DeiT-Base-Distilled is given by Equation~\eqref{eq:deit_loss}:

\begin{equation} \label{eq:deit_loss}
L_{\text{total}} = \alpha L_{\text{CE}} + (1 - \alpha) L_{\text{KD}},
\end{equation}

where $L_{\text{CE}}$ represents the cross-entropy loss computed against the true labels, $L_{\text{KD}}$ is the knowledge distillation loss obtained from the teacher model’s predictions, and $\alpha$ is a hyperparameter that balances the contributions of the two loss components. By optimizing this combined loss function, Equation~\eqref{eq:deit_loss}, DeiT-Base-Distilled learns to leverage both hard labels and soft labels, which helps it generalize well even on small datasets.

\textbf{Regularization and augmentation}

DeiT-Base-Distilled also employs strong augmentation techniques such as Mixup, CutMix, and random erasing to introduce variability in the training data and prevent overfitting \cite{deittouvron2021training}. Mixup generates hybrid samples by combining pairs of images and their labels, creating smoother transitions between classes and reducing the risk of overfitting \cite{zhang2017mixup}. CutMix cuts and pastes patches between images, proportionally mixing the labels based on the area of the patches, which retains more information and improves model robustness~\cite{yun2019cutmix}. Random Erasing randomly removes rectangular regions of an image during training, simulating occlusions and making the model more resilient to missing or corrupted information \cite{zhong2020random}.


\subsection{Textual feature extraction utilizing DeBERTa-base}

DeBERTa-base (Decoding-enhanced BERT with Disentangled Attention) is a powerful transformer model specifically designed to capture and extract meaningful textual features~\cite{debertahe2020deberta}. In the DM-WAT framework, DeBERTa-base effectively processes clinical notes, which are often complex and contain domain-specific language, enabling more accurate decision-making for wound care. The model incorporates two primary architectural innovations—disentangled attention and an enhanced mask decoder—tailored to improve contextual understanding, particularly when applied to small or specialized datasets, such as clinical notes.

\textbf{Disentangled Attention}

\begin{figure}
\centering
\includegraphics[width=0.4\textwidth]{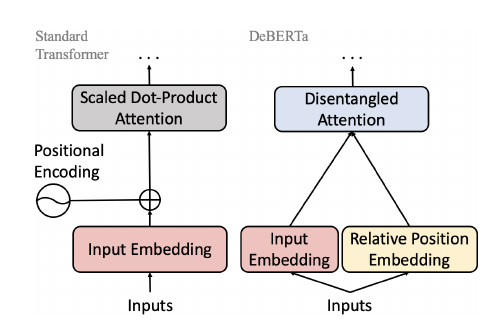}
\caption{Comparison of Standard BERT Attention (Left) and DeBERTa's Disentangled Attention (Right) \cite{qian2022limitations}. DeBERTa separates token embeddings from position embeddings, allowing it to compute more precise attention scores based on both content and relative position.}
\label{fig:disentangled_attention}
\end{figure}

One of DeBERTa’s key innovations is disentangled attention, where each token is represented by two separate embeddings: one for semantic content and one for positional context \cite{debertahe2020deberta}. This separation allows DeBERTa to compute more precise attention scores based on both the content of each token and its relative position, which enhances the model's ability to capture complex relationships within clinical notes. Figure~\ref{fig:disentangled_attention} provides an illustration of how DeBERTa employs separate embeddings for text content and position. 

In DeBERTa, the attention score for each token is computed by combining the content embedding ($C$) and positional embedding ($P$) for each query and key, as shown in Equation~\eqref{eq:deberta_attention}:

\begin{multline} \label{eq:deberta_attention}
\text{Attention}(Q_C + Q_P, K_C + K_P, V) = \\
\text{softmax}\left(\frac{(Q_C + Q_P)(K_C + K_P)^T}{\sqrt{d_k}}\right) \times V,
\end{multline}

where $Q_C$ and $K_C$ represent the content embeddings, $Q_P$ and $K_P$ are the positional embeddings for the query and key, respectively, and $d_k$ is the scaling factor. Equation~\eqref{eq:deberta_attention} highlights how DeBERTa separately processes content and positional information to compute attention scores, enabling it to better capture the meaning of each word as well as its structural role within a sentence. This feature is particularly valuable for understanding nuanced medical language in clinical notes.

\textbf{Enhanced mask decoder}

\begin{figure}
\centering
\includegraphics[width=0.5\textwidth]{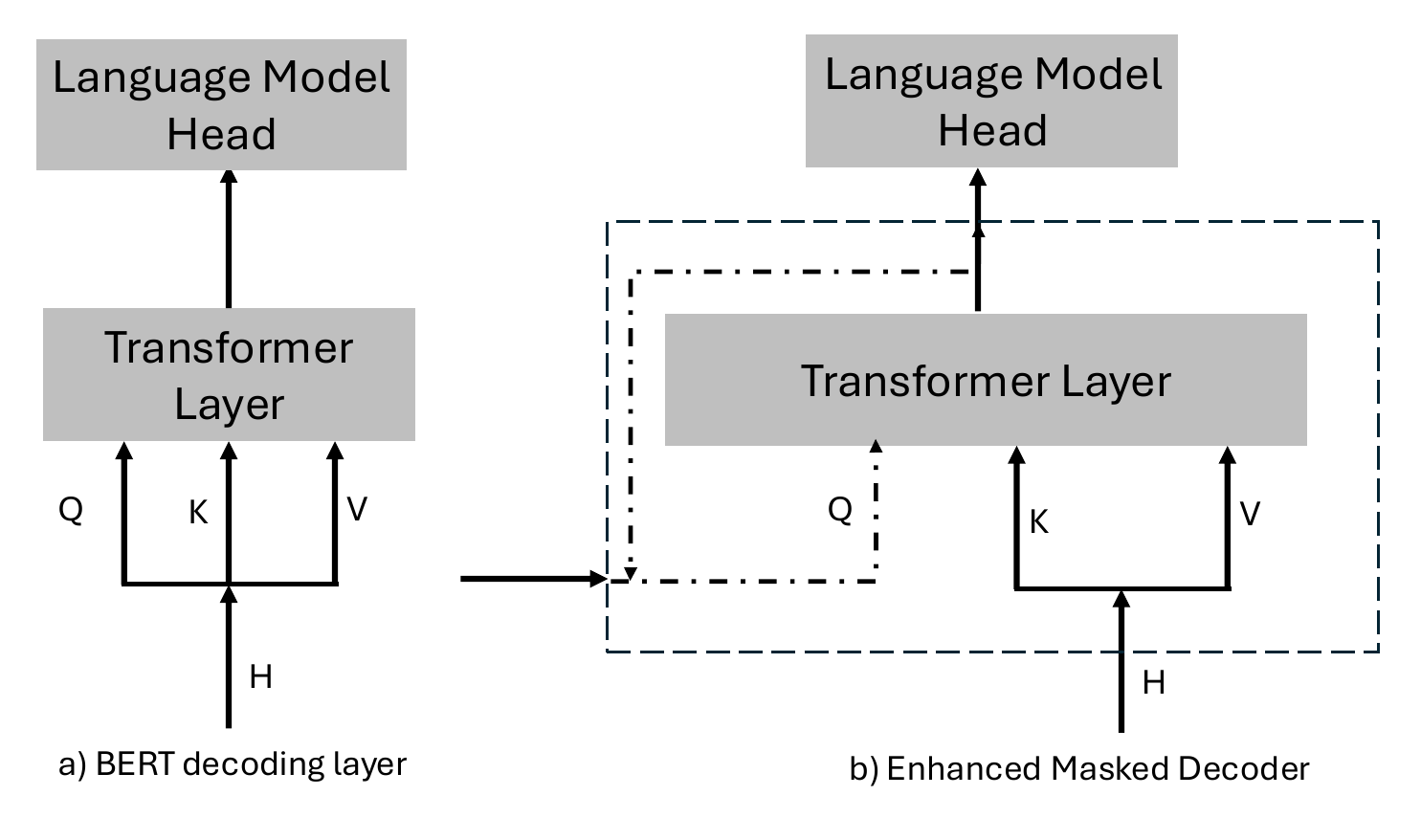}
\caption{Comparison of BERT's Decoder (Left) and DeBERTa's Enhanced Mask Decoder (Right). DeBERTa’s EMD integrates absolute position embeddings, making it more flexible and capable of capturing complex relationships in clinical text \cite{debertahe2020deberta}.}
\label{fig:enhanced_mask_decoder}
\end{figure}

In addition to disentangled attention, DeBERTa features an Enhanced Mask Decoder (EMD), which further improves its interpretive ability by integrating absolute position embeddings directly into the decoding process \cite{debertahe2020deberta}. This differs from standard BERT, where only the hidden states from the previous layer are used in decoding. DeBERTa’s EMD enables more flexible and accurate decoding by allowing the model to incorporate different input types, including hidden states and absolute positions, into its final predictions. Figure \ref{fig:enhanced_mask_decoder} compares the standard BERT decoder with DeBERTa’s enhanced mask decoder.
This enhanced decoding process allows DeBERTa to better capture relationships between clinical terms and improve token prediction accuracy, which is essential for understanding and processing the nuanced language of clinical notes.

\textbf{Performance  on small data}
The architecture of DeBERTa-base, incorporating disentangled attention and an enhanced mask decoder, makes it well-suited for handling limited and complex datasets, such as clinical notes in wound care settings. These features allow DeBERTa to capture nuanced dependencies, even when working with a relatively small dataset. Consequently, it can extract meaningful insights from clinical notes, making it a strong candidate for the DM-WAT framework, where reliable and interpretable analysis of limited medical text data is critical for informed decision-making.

\subsection{Multimodal fusion}

\begin{figure}
\centering
\includegraphics[width=0.4\textwidth]{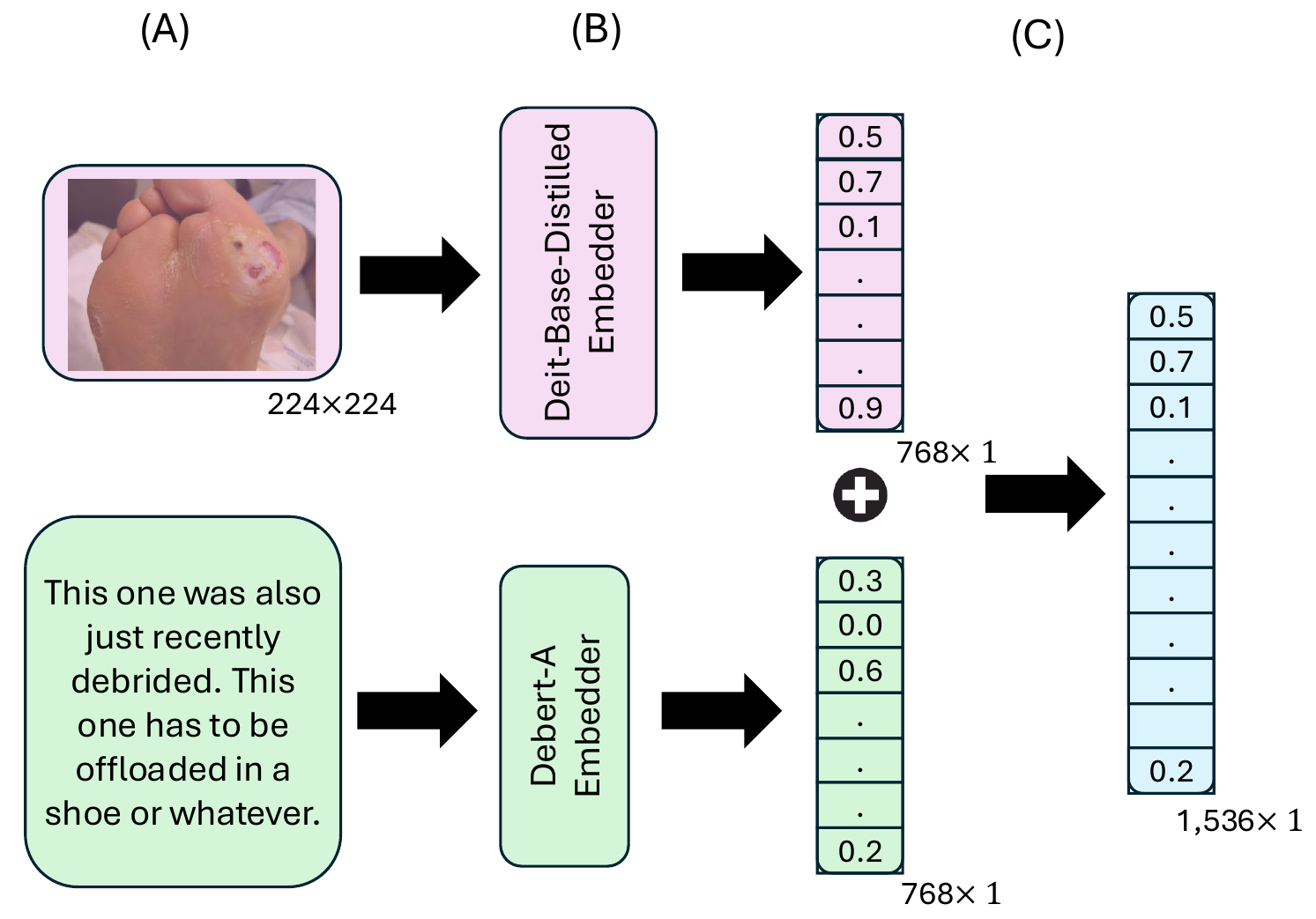}
\caption{Illustration of the Intermediate Fusion Process: (A) Input modalities (image and text), (B) modality-specific embedders (DeiT-Base-Distilled for images, DeBERTa-base for text), (C) extracted embedding vectors concatenated into a 1,536-dimensional combined representation.}
\label{fig:intermediate_fusion_example}
\end{figure}

DM-WAT combines image and text features utilizing intermediate fusion, integrating high-level representations from DeiT-Base-Distilled and DeBERTa-base. This fusion strategy preserves unique modality-specific features while enabling a more comprehensive understanding of the data. The combined features form a unified vector representation, which is fed into a classifier \cite{intermediateFusionyoo2019deep}. Figure \ref{fig:intermediate_fusion_example} demonstrates the fusion process.

\subsection{Support Vector Machines (SVM) to classify fused representation}
The fused features are classified using a Support Vector Machine (SVM), an algorithm that identifies optimal hyperplanes to separate data points across multiple classes. SVM works well on limited datasets and offers clear decision boundaries. The classification is performed using the decision function, as expressed by  Equation~\eqref{eq:decision_function}:

\begin{equation} \label{eq:decision_function}
F(x) = Wx + b,
\end{equation}

where $W$ represents the weight vector and $b$ is the bias term. The margin, defined as the distance between the hyperplane and the nearest support vectors, is maximized to achieve optimal separation. This optimization can be expressed as:

\begin{equation} \label{eq:margin_max}
\max \left( \frac{2}{\lVert W \rVert} \right).
\end{equation}

Equation~\eqref{eq:margin_max} ensures that the model maximizes the separation between classes, enhancing the robustness of predictions.

\subsection{Interpretation algorithms} \label{secInterpretation}

The Score-CAM (for images) and Captum (for text) interpretability algorithms provide insight into model predictions, enhancing the model's transparency and trustworthiness.

\subsubsection{Score-CAM}
Score-CAM interprets model predictions by generating heatmaps that highlight predictive regions in wound images. It creates these maps by masking different parts of the image and observing the changes in model confidence. The final Score-CAM heatmap is calculated using Equation~\eqref{eq:score_cam}:

\begin{equation} \label{eq:score_cam}
L^{c}_{\text{Score-CAM}} = \text{ReLU} \left( \sum_k \alpha^c_k A^k_l \right),
\end{equation}

where $\alpha^c_k$ represents the normalized importance of each activation map $A^k_l$. Equation~\eqref{eq:score_cam} ensures that only the positive contributions of the activation maps are considered, focusing the heatmap on regions most relevant to the model's prediction.

\subsubsection{Captum}

Captum utilizes Integrated Gradients to interpret textual predictions, identifying predictive words in clinical notes. Integrated Gradients compute the attribution score as shown in Equation~\eqref{eq:integrated_gradients}:

\begin{equation} \label{eq:integrated_gradients}
\text{IntegratedGrad}(x) = (x - x') \times \int_{\alpha=0}^{1} \frac{\partial f(x' + \alpha \times (x - x'))}{\partial x} d\alpha,
\end{equation}

where $x$ is the actual input, $x'$ is a baseline input (representing a neutral or reference point for the input), and $f$ is the model’s output function. Equation~\eqref{eq:integrated_gradients} highlights tokens in clinical notes that significantly impact the model’s predictions by evaluating the path integral of gradients between the baseline and the actual input. This enables clinicians to understand the reasoning behind each referral decision.

\section{Evaluation and results} \label{sec:evaluation}

This section presents the evaluation metrics, results of different types of models (image-only, text-only, and multimodal), and interpretation algorithms used to analyze the performance of DM-WAT. Each result is analyzed and discussed to reason about factors contributing to model performance. 

\subsection{Evaluation metrics}

To assess the performance of DM-WAT and baseline models, the following standard metrics were utilized:

\textbf{Accuracy} measures the overall correctness of predictions, as defined in Equation~\eqref{eq:accuracy}:

\begin{equation} \label{eq:accuracy}
\text{Accuracy} = \frac{TP + TN}{TP + TN + FP + FN},
\end{equation}

where $TP$ is True Positives, $TN$ is True Negatives, $FP$ is False Positives, and $FN$ is False Negatives. 

\textbf{Precision} evaluates the ratio of correctly predicted positives to all predicted positives, as shown in Equation~\eqref{eq:precision}:

\begin{equation} \label{eq:precision}
\text{Precision} = \frac{TP}{TP + FP}.
\end{equation}

\textbf{Recall} measures the ratio of correctly predicted positives to all actual positives, as defined in Equation~\eqref{eq:recall}:

\begin{equation} \label{eq:recall}
\text{Recall} = \frac{TP}{TP + FN}.
\end{equation}

\textbf{F1 Score} balances Precision and Recall, as shown in Equation~\eqref{eq:f1}:

\begin{equation} \label{eq:f1}
\text{F1} = 2 \times \frac{\text{Precision} \times \text{Recall}}{\text{Precision} + \text{Recall}}.
\end{equation}

\subsection{Results}

Stratified 5-fold cross-validation was employed to ensure consistent class distribution across folds. This method reduces bias caused by imbalanced datasets, facilitating reliable evaluations.

\subsubsection{Image-Only classifier evaluation}

The image-only classifiers were evaluated using both non-augmented and augmented datasets (Table~\ref{tab:class_distribution}).

\begin{table}
\centering
\caption{Number of Images in Each Class for Non-Augmented and Augmented Datasets}
\begin{tabular}{|c|c|c|}
\hline
\textbf{Class} & \textbf{Without Augmentation} & \textbf{With Augmentation} \\ \hline
Class 1        & 26                             & 1950                        \\ \hline
Class 2        & 40                             & 1850                        \\ \hline
Class 3        & 139                            & 2085                        \\ \hline
\end{tabular}
\label{tab:class_distribution}
\end{table}

The DeiT-Base-Distilled model achieved the highest F1 score of 67\% with augmentation (Table~\ref{tab:image_results}). Vision Transformers (ViTs) such as DeiT excel in capturing long-range dependencies via self-attention, making them well-suited for recognizing subtle patterns in wound images. Additionally, knowledge distillation allows the model to generalize effectively, even on small datasets, by learning from a teacher model that emphasizes clinically relevant features.

 Augmentation significantly improved performance, increasing the F1 score from 33\% without augmentation to 67\% with augmentation. By introducing variations such as rotation, flipping, and cropping, augmentation exposes the model to diverse scenarios, helping it focus on invariant features rather than overfitting. This is especially important for small, imbalanced datasets like ours, where augmentation not only increases dataset size but also balances class distributions (Table~\ref{tab:class_distribution}). The combination of augmentation and DeiT's architecture enhances robustness and accuracy, making it the best performing model in this study.

\begin{table*}
\centering
\caption{Performance of CNN and ViT-based Models on Wound Image Classification}
\small 
\begin{tabular}{|l|c|c|c|c|c|c|c|c|}
\hline
\multirow{2}{*}{\textbf{Model}} & \multicolumn{4}{c|}{\textbf{Without Augmentation}} & \multicolumn{4}{c|}{\textbf{With Augmentation}} \\ \cline{2-9}
                                & \textbf{Acc} & \textbf{Rec} & \textbf{Prec} & \textbf{F1} & \textbf{Acc} & \textbf{Rec} & \textbf{Prec} & \textbf{F1} \\ \hline
\multicolumn{9}{|c|}{\textbf{CNN-based models}} \\ \hline
VGG16                           & 53$\pm$7     & 42$\pm$7     & 33$\pm$5      & 37$\pm$6    & 68$\pm$2     & 30$\pm$3     & 28$\pm$2      & 28$\pm$2    \\ \hline
ResNet50                        & 55$\pm$4     & 34$\pm$5     & 32$\pm$5      & 33$\pm$3    & 67$\pm$1     & 33$\pm$2     & 24$\pm$1      & 27$\pm$1    \\ \hline
EfficientNetB0                  & 59$\pm$4     & 39$\pm$3     & 38$\pm$4      & 39$\pm$3    & 66$\pm$3     & 34$\pm$3     & 31$\pm$4      & 31$\pm$4    \\ \hline
MobileNetV2                     & 60$\pm$5     & 43$\pm$6     & 39$\pm$4      & 40$\pm$4    & 68$\pm$1     & 33$\pm$1     & 25$\pm$2      & 27$\pm$1    \\ \hline
\multicolumn{9}{|c|}{\textbf{ViT-based models}} \\ \hline
DeiT-Tiny                       & 67$\pm$5     & 39$\pm$4     & 37$\pm$4      & 39$\pm$4    & 67$\pm$4     & 67$\pm$4     & 65$\pm$6      & 66$\pm$4    \\ \hline
\textbf{DeiT-Base-Distilled}    & 66$\pm$5     & 33$\pm$2     & 34$\pm$5      & 33$\pm$5    & \textbf{70$\pm$4} & \textbf{72$\pm$3} & \textbf{66$\pm$5} & \textbf{67$\pm$5} \\ \hline
ViT-Hybrid-Base                 & 68$\pm$6     & 40$\pm$6     & 37$\pm$4      & 38$\pm$5    & 69$\pm$4     & 70$\pm$2     & 64$\pm$3      & 65$\pm$3    \\ \hline
\end{tabular}
\label{tab:image_results}
\end{table*}


\subsubsection{Text-only classifier evaluation}

Table~\ref{tab:text_results} presents the performance of BERT-based models for wound text classification. Among the models evaluated, \textbf{DeBERTa-base} achieved the highest F1 score (65\%) with augmented data. DeBERTa's superior performance can be attributed to its \textit{disentangled attention mechanism}, which separates content and positional embeddings, allowing for more nuanced understanding of word relationships. Additionally, its \textit{enhanced mask decoder} improves the model's ability to capture complex dependencies within clinical notes. These innovations make DeBERTa especially well-suited for interpreting domain-specific language with limited and contextually dense datasets, such as clinical wound descriptions.

\begin{table*}[!t]
\centering
\caption{Performance of BERT-based models on wound text classification with and without augmentation}
\small
\begin{tabular}{|l|c|c|c|c|c|c|c|c|}
\hline
\multirow{2}{*}{\textbf{Model}} & \multicolumn{4}{c|}{\textbf{Without Augmentation}} & \multicolumn{4}{c|}{\textbf{With Augmentation }} \\ \cline{2-9}
                                & \textbf{Acc} & \textbf{Rec} & \textbf{Prec} & \textbf{F1} & \textbf{Acc} & \textbf{Rec} & \textbf{Prec} & \textbf{F1} \\ \hline
\multicolumn{9}{|c|}{\textbf{BERT-based models}} \\ \hline
Med-BERT                        & 64$\pm$3        & 62$\pm$2      & 56$\pm$3        & 61$\pm$3  & 65$\pm$3        & 62$\pm$2      & 55$\pm$3        & 60$\pm$3  \\ \hline
Dr-BERT                         & 65$\pm$5        & 62$\pm$6      & 55$\pm$6        & 58$\pm$6  & 65$\pm$3        & 61$\pm$4      & 56$\pm$2        & 58$\pm$4  \\ \hline
BERT-base-uncased               & 70$\pm$2        & 68$\pm$1      & 60$\pm$3        & 63$\pm$3  & 68$\pm$5        & 69$\pm$3      & \textbf{61$\pm$5}        & \textbf{66$\pm$5}  \\ \hline
BioBERT-v1.1                    & 68$\pm$3        & 61$\pm$4      & 51$\pm$4        & 55$\pm$4  & 69$\pm$2        & 62$\pm$3      & 52$\pm$4        & 55$\pm$4  \\ \hline
RoBERTa-base                    & 70$\pm$4        & 66$\pm$5      & 58$\pm$4        & 60$\pm$4  & 70$\pm$6        & 66$\pm$5      & 58$\pm$6        & 61$\pm$6  \\ \hline
\textbf{DeBERTa-base}                    & {73$\pm$4} & 69$\pm$3      & 60$\pm$2        & 63$\pm$3  & \textbf{74$\pm$4} & \textbf{70$\pm$6}      & \textbf{61$\pm$4}        & 65$\pm$5  \\ \hline
\multicolumn{9}{|c|}{\textbf{Baseline Model}} \\ \hline
Nguyen et al. (2020) HAN SMOTE  & 64$\pm$5        & 61$\pm$3      & 51$\pm$5        & 53$\pm$4  & 63$\pm$4                & 61$\pm$3        & 52$\pm$4                 & 53$\pm$4         \\ \hline
\end{tabular}%
\label{tab:text_results}
\end{table*}

Despite augmenting textual data with GPT-generated descriptions, the overall improvement in performance was minimal across all models. While the augmented text provided additional variability, its utility was constrained by potential mismatches in the quality or relevance of the synthetic text compared to expert-provided clinical notes. For \textbf{DeBERTa-base}, the F1 score increased marginally from 63\% (without augmentation) to 65\% (with augmentation), indicating that the added text may have introduced only slight benefits in feature diversity but did not significantly improve the model's generalization. This suggests that while GPT-generated text can complement datasets with limited expert annotations, it may not fully replicate the depth and precision of expert-authored clinical notes required for optimal model performance.

\subsubsection{Multimodal classifier evaluation}

This evaluation aims to assess the performance of intermediate fusion of image and text data in making referral decisions. For the DM-WAT model, we selected the best-performing classifiers for feature extraction—DeiT-Base-Distilled for vision features and DeBERTa-Base for text features. These features were then combined using intermediate fusion, with classification by either SVM or MLP. All models were trained and tested on augmented versions of the dataset. The DM-WAT algorithm, using both SVM and MLP classifiers, was trained for 20 epochs with a learning rate of 1e-6.

\begin{table*}[!t]
\centering
\caption{Comparison of Best Image, Text, and Multimodal Models on Wound Classification}
\small
\begin{tabular}{|l|c|c|c|c|}
\hline
\textbf{} & \textbf{Accuracy} & \textbf{Recall} & \textbf{Precision} & \textbf{F1} \\ \hline
\multicolumn{5}{|c|}{\textbf{Image-based}} \\ \hline
Deit-tiny                    & 67$\pm$4  & 67$\pm$4   & 65$\pm$6   & 66$\pm$4 \\ \hline
Deit-base-distilled          & 70$\pm$4  & 72$\pm$3   & 66$\pm$5   & 67$\pm$5 \\ \hline
Vit-hybrid-base              & 69$\pm$4  & 70$\pm$2   & 64$\pm$3   & 65$\pm$3 \\ \hline
\multicolumn{5}{|c|}{\textbf{Text-based}} \\ \hline
RoBERTa-base                 & 70$\pm$6  & 66$\pm$5   & 58$\pm$6   & 61$\pm$6 \\ \hline
DeBERTa-base                 & 74$\pm$4  & 70$\pm$6   & 61$\pm$4   & 65$\pm$5 \\ \hline
Nguyen et al. (2020) HAN SMOTE & 64$\pm$3  & 65$\pm$4   & 60$\pm$3   & 62$\pm$4 \\ \hline
\multicolumn{5}{|c|}{\textbf{Multimodal}} \\ \hline
DM-WAT (DeBERTa + DeiT + MLP)  & 76$\pm$4  & 72$\pm$5   & \textbf{67$\pm$3}   & 69$\pm$3 \\ \hline
\textbf{DM-WAT (DeBERTa + DeiT + SVM)}  & \textbf{77$\pm$3}  & \textbf{73$\pm$3}   & \textbf{67$\pm$2}   & \textbf{70$\pm$2} \\ \hline
Nguyen et al. (2020) Previous Multimodal SOTA & 71$\pm$6  & 64$\pm$4   & 54$\pm$6   & 61$\pm$4 \\ \hline
\end{tabular}%

\label{tab:multimodal_results}
\end{table*}

As shown in Table \ref{tab:multimodal_results}, DM-WAT outperformed both single-modality models and Nguyen's multimodal baseline. This success is due to the advanced feature extractors for both image and text data, which provided a more accurate view of wound characteristics. The use of intermediate fusion effectively combined these features, improving robustness and accuracy. Additionally, the multimodal results suggest that DM-WAT with the SVM classifier achieved slightly better results than the MLP classifier. Overall, these findings demonstrate that utilizing both image and text data yields more accurate wound referral decisions than relying on a single type of data.

\subsection{Interpretation of model decisions}

After evaluating DM-WAT's performance, we wanted to understand how the models interpret inputs and identify key parts of input data that are predictive of the target labels. As mentioned in Section \ref{secInterpretation},  Score-CAM and Captum are utilized to interpret image and text inputs to the DeiT-Base-Distilled and DeBERTa-Base models respectively. These interpretation algorithms were applied to all original images, and for text input,  GPT-4-generated text was utilized because it corresponds better to the image details and provides more useful information.

\begin{figure*}[htbp]
\centering
\includegraphics[width=\textwidth]{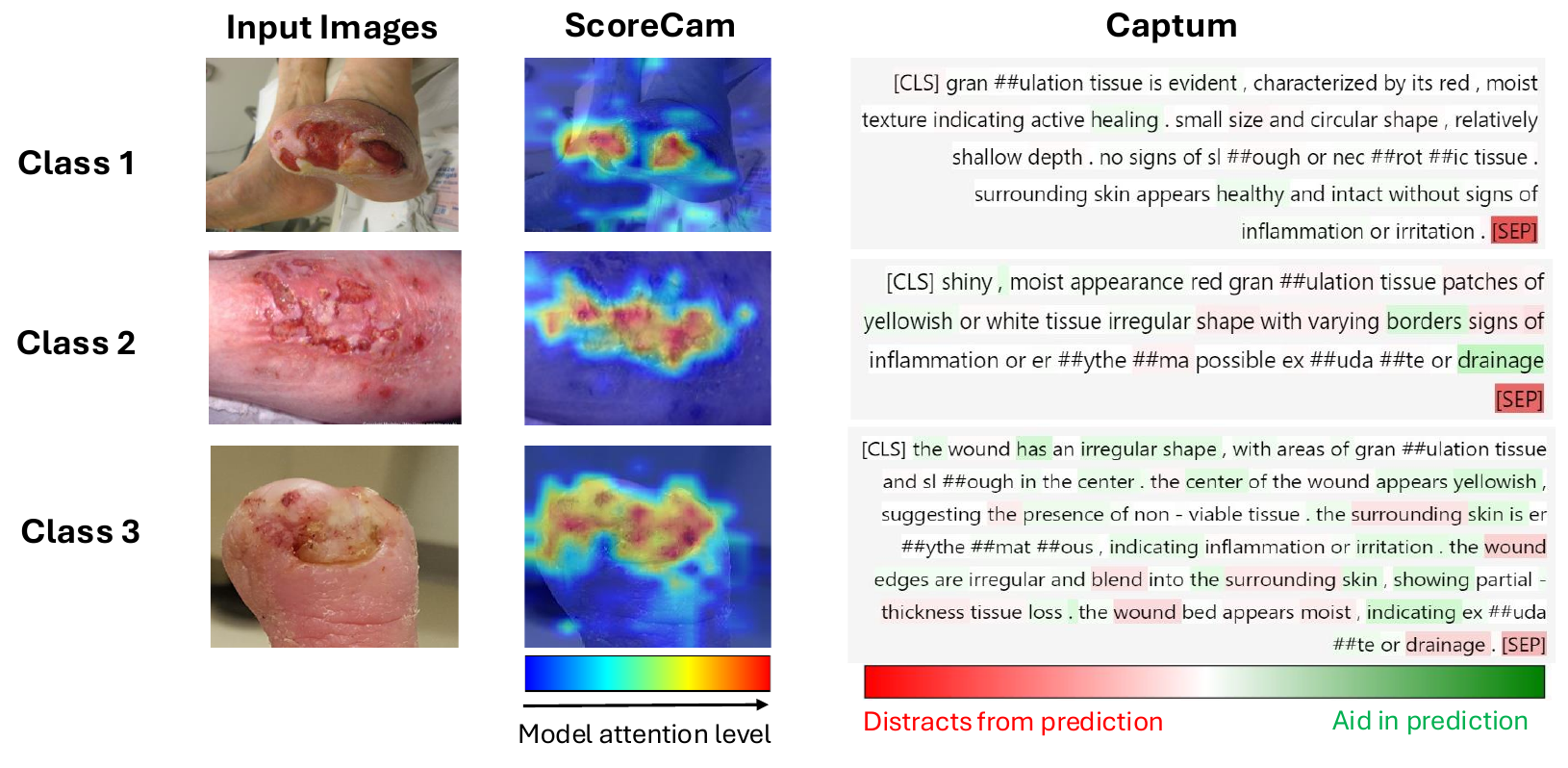}
\caption{Model interpretations using Score-CAM for images (left) and Captum for clinical notes (right). In Score-CAM, red indicates high attention, and blue indicates low attention of the vision model for classification. In Captum, green text aids predictions, while red text distracts from predictions. Examples representing the three wound care decision classes are shown: Class 1 (continue treatment), Class 2 (non-urgent referral), and Class 3 (urgent referral).}

\label{fig:interpretation}
\end{figure*}
As shown in Figure \ref{fig:interpretation}, both Score-CAM and Captum generally focused on the parts of the input that are significant for making predictions. Interestingly, they often highlight the same areas across different modalities. For instance, in the Class 3 example, both models pay attention to the irregular shape and the yellow areas—Score-CAM highlights the yellow regions in the image, while Captum emphasizes the yellowish description in the text. This overlap suggests that the models are consistently identifying the most relevant features in both image and text data, reinforcing the effectiveness of using both data modalities for robust prediction.

\section{Discussion} \label{sec:discussion}

\textbf{The scarcity of labeled data and class imbalance were effectivly addressed using data augmentation and transfer learning.} Data augmentation for images and pre-trained models such as DeiT and DeBERTa generally improved results and allowed the model to perform effectively despite limited data availability. For missing and contradictory clinical notes, GPT-4 was employed to generate supplementary text, enriching the dataset with alternative perspectives. For inconsistent referral decisions from wound experts, a conservative approach was adopted, prioritizing patient safety by selecting the more urgent recommendation. Stratified cross-fold validation was implemented to ensure proportional representation of classes, enhancing the model's ability to generalize across data subsets.

\textbf{Multimodal fusion significantly improved DM-WAT's referral decision accuracy.} The results highlighted that the multimodal approach achieved 77\% accuracy and an F1 score of 70\%, outperforming single-modal models. Vision Transformer (ViT) models, particularly DeiT-Base-Distilled, excelled due to their self-attention mechanisms, achieving 70\% accuracy and 67\% F1 score. Similarly, DeBERTa demonstrated superior text classification performance, achieving 74\% accuracy and 65\% F1 score, attributed to its advanced attention mechanisms and enhanced mask decoder. Data augmentation and majority voting further improved model robustness, with DeiT-Base-Distilled’s F1 score increasing from 42\% to 69\%.

\textbf{Interpretability methods provided insights into the model's decision-making}. These included Captum for text and Score-CAM for images, potentially increasing clinician trust in model outputs and the likelihood of adoption. 

\textbf{Unexpected findings revealed limitations and areas for improvement.} The impact of text augmentation was limited, suggesting that either the model had reached its optimal performance or the augmented text lacked sufficient diversity. Slight overfitting persisted despite regularization techniques, indicating the need for advanced data expansion strategies. A key limitation of this study is its reliance on a single dataset with unknown demographics and wound types that were not specified, which may limit generalizability. Future work should specify wound types and investigate differences in results on different wound types and include diverse datasets to validate the robustness and adaptability of DM-WAT across varied patient populations.

\section{Conclusion and future work} \label{sec:conclusion}

\subsection{Conclusion}

This paper introduced DM-WAT, a novel machine learning framework designed to assist clinicians in wound assessment and referral decision-making. The proposed system integrates multimodal data, combining wound images and clinical notes, to provide a comprehensive analysis of wound characteristics. The multimodal DM-WAT model, which employed DeBERTa for text and DeiT-Base-Distilled for images, demonstrated superior performance, achieving an accuracy of 77\% and an F1 score of 70\%. 

The success of DM-WAT is attributed to several key contributions. Transfer learning enabled advanced models such as DeiT-Base-Distilled and DeBERTa to leverage pre-trained knowledge, enhancing performance on a limited dataset. DeiT-Base-Distilled emerged as the best-performing image-based model, achieving an F1 score of 67\%, while DeBERTa led text-based models with an F1 score of 65\%. Data augmentation, particularly for images, and majority voting significantly improved model robustness and reduced overfitting. For instance, the application of augmentation and majority voting increased the F1 score of DeiT-Base-Distilled from 42\% to 69\% on the test set.

Additionally, the Score-CAM and Captum interpretability methods provided visual and textual highlighted regions of input data that were predictive of the referral target labels, facilitating sensemaking and could potentially aid non-specialist clinicians in understanding the model's outputs. Overall, DM-WAT represents a significant advance in improving wound care decision-making and patient outcomes for chronic wound management.

\subsection{Future work}

Although DM-WAT has shown promising results, several avenues for improvement and further research remain. First, the current approach to text data augmentation using GPT-4 had limited impact. Exploring advanced prompt engineering techniques, such as few-shot and chain-of-thought prompting, could improve the quality and diversity of generated text, potentially enhancing model performance.

For image augmentation, deep learning-based methods such as diffusion models or Generative Adversarial Networks (GANs) could provide more realistic and diverse synthetic data, further improving the model's generalization capabilities. Advanced fusion strategies, including attention-based fusion, could be explored to better combine visual and textual features, improving robustness and accuracy.

Incorporating wound assessment scores such the PWAT score auto-generated by external models during intermediate fusion could enhance decision-making by providing additional wound information. To build trust with non-specialist users, the system could provide explanations and justifications for its decisions in a format tailored for clinicians, further increasing their understanding and trust the recommendations.

Given the limited dataset, semi-supervised learning methods, such as Semi-Supervised Progressive Multi-Granularity (SS-PMG)~\cite{liu2023chronic} training, could be employed to leverage additional unlabeled wound data. Reinforcement learning with human feedback (RLHF) could address inconsistencies in expert referrals by integrating feedback to refine the model's decision-making, dynamically prioritizing more reliable expert input.

Finally, expanding the dataset with more labeled data from experts will improve DM-WAT’s accuracy and generalizability, reliability and utility, ensuring its applicability across diverse clinical scenarios.


%



\section*{Acknowledgments}
This research was performed using computational resources supported by the Academic \& Research Computing group at Worcester Polytechnic Institute. Additionally, this work was supported by the National Institutes of Health (NIH) under grant number 1R01EB031910-01A1, titled "Smartphone-based wound infection screener by combining thermal images and photographs using deep learning methods."


\ifCLASSOPTIONcaptionsoff
  \newpage
\fi



%


\bibliographystyle{IEEEtran}

\bibliography{main} 

\begin{thebibliography}{10}
\providecommand{\url}[1]{#1}
\csname url@samestyle\endcsname
\providecommand{\newblock}{\relax}
\providecommand{\bibinfo}[2]{#2}
\providecommand{\BIBentrySTDinterwordspacing}{\spaceskip=0pt\relax}
\providecommand{\BIBentryALTinterwordstretchfactor}{4}
\providecommand{\BIBentryALTinterwordspacing}{\spaceskip=\fontdimen2\font plus
\BIBentryALTinterwordstretchfactor\fontdimen3\font minus \fontdimen4\font\relax}
\providecommand{\BIBforeignlanguage}[2]{{%
\expandafter\ifx\csname l@#1\endcsname\relax
\typeout{** WARNING: IEEEtran.bst: No hyphenation pattern has been}%
\typeout{** loaded for the language `#1'. Using the pattern for}%
\typeout{** the default language instead.}%
\else
\language=\csname l@#1\endcsname
\fi
#2}}
\providecommand{\BIBdecl}{\relax}
\BIBdecl

\bibitem{jarbrink2016prevalence}
K.~J{\"a}rbrink, G.~Ni, H.~S{\"o}nnergren, A.~Schmidtchen, C.~Pang, R.~Bajpai, and J.~Car, ``Prevalence and incidence of chronic wounds and related complications: a protocol for a systematic review,'' \emph{Systematic reviews}, vol.~5, pp. 1--6, 2016.

\bibitem{sen2021human}
C.~K. Sen, ``Human wound and its burden: updated 2020 compendium of estimates,'' \emph{Advances in wound care}, vol.~10, no.~5, pp. 281--292, 2021.

\bibitem{1frykbergrobert2015challenges}
G.~FrykbergRobert \emph{et~al.}, ``Challenges in the treatment of chronic wounds,'' \emph{Advances in wound care}, 2015.

\bibitem{2sen2019human}
C.~K. Sen, ``Human wounds and its burden: an updated compendium of estimates,'' pp. 39--48, 2019.

\bibitem{4nguyen2020machine}
H.~Nguyen, E.~Agu, B.~Tulu, D.~Strong, H.~Mombini, P.~Pedersen, C.~Lindsay, R.~Dunn, and L.~Loretz, ``Machine learning models for synthesizing actionable care decisions on lower extremity wounds,'' \emph{Smart Health}, vol.~18, p. 100139, 2020.

\bibitem{5jarbrink2017humanistic}
K.~J{\"a}rbrink, G.~Ni, H.~S{\"o}nnergren, A.~Schmidtchen, C.~Pang, R.~Bajpai, and J.~Car, ``The humanistic and economic burden of chronic wounds: a protocol for a systematic review,'' \emph{Sys. reviews}, vol.~6, pp. 1--7, 2017.

\bibitem{WAT2hon2010prospective}
J.~Hon, K.~Lagden, A.-M. McLaren, D.~O'Sullivan, L.~Orr, P.~E. Houghton, and M.~G. Woodbury, ``A prospective, multicenter study to validate use of the push in patients with diabetic, venous, and pressure ulcers.'' \emph{Ostomy/wound management}, vol.~56, no.~2, pp. 26--36, 2010.

\bibitem{PWATthompson2013reliability}
N.~Thompson, L.~Gordey, H.~Bowles, N.~Parslow, and P.~Houghton, ``Reliability and validity of the revised photographic wound assessment tool on digital images taken of various types of chronic wounds,'' \emph{Advances in skin \& wound care}, vol.~26, no.~8, pp. 360--373, 2013.

\bibitem{cnn1zhang2022survey}
R.~Zhang, D.~Tian, D.~Xu, W.~Qian, and Y.~Yao, ``A survey of wound image analysis using deep learning: Classification, detection, and segmentation,'' \emph{IEEE Access}, vol.~10, pp. 79\,502--79\,515, 2022.

\bibitem{cnn2rostami2021multiclass}
B.~Rostami, D.~Anisuzzaman, C.~Wang, S.~Gopalakrishnan, J.~Niezgoda, and Z.~Yu, ``Multiclass wound image classification using an ensemble deep cnn-based classifier,'' \emph{Computers in Biology and Medicine}, vol. 134, p. 104536, 2021.

\bibitem{vitdosovitskiy2020image}
A.~Dosovitskiy, L.~Beyer, A.~Kolesnikov, D.~Weissenborn, X.~Zhai, T.~Unterthiner, M.~Dehghani, M.~Minderer, G.~Heigold, S.~Gelly \emph{et~al.}, ``An image is worth 16x16 words: Transformers for image recognition at scale,'' \emph{arXiv preprint arXiv:2010.11929}, 2020.

\bibitem{vitApp1mohan2024vision}
R.~Mohan, N.~S.~M. Raja, R.~Damasevicius, D.~Taniar, S.~Prabha, and V.~Rajinikanth, ``Vision transformer based diabetic foot-ulcer detection: A study,'' in \emph{Int'l Conf. Bio Signals, Images, and Instrumentation (ICBSII)}.\hskip 1em plus 0.5em minus 0.4em\relax IEEE, 2024, pp. 1--4.

\bibitem{tfidf1lasko2013computational}
T.~A. Lasko, J.~C. Denny, and M.~A. Levy, ``Computational phenotype discovery using unsupervised feature learning over noisy, sparse, and irregular clinical data,'' \emph{PloS one}, vol.~8, no.~6, p. e66341, 2013.

\bibitem{bertdevlin2018bert}
J.~Devlin, ``Bert: Pre-training of deep bidirectional transformers for language understanding,'' \emph{arXiv preprint arXiv:1810.04805}, 2018.

\bibitem{yildirim2024multimodal}
N.~Yildirim, H.~Richardson, M.~T. Wetscherek, J.~Bajwa, J.~Jacob, M.~A. Pinnock, S.~Harris, D.~Coelho De~Castro, S.~Bannur, S.~Hyland \emph{et~al.}, ``Multimodal healthcare ai: identifying and designing clinically relevant vision-language applications for radiology,'' in \emph{Proceedings of the CHI Conference on Human Factors in Computing Systems}, 2024, pp. 1--22.

\bibitem{hartsock2024vision}
I.~Hartsock and G.~Rasool, ``Vision-language models for medical report generation and visual question answering: A review,'' \emph{Frontiers in Artificial Intelligence}, vol.~7, p. 1430984, 2024.

\bibitem{16mombini2020design}
H.~Mombini, B.~Tulu, D.~Strong, E.~Agu, H.~Nguyen, C.~Lindsay, L.~Loretz, P.~Pedersen, and R.~Dunn, ``Design of a machine learning system for prediction of chronic wound management decisions,'' in \emph{Designing for Digital Transformation. Co-Creating Services with Citizens and Industry: 15th Int'l Conf. Design Science Research in Info Sys. and Tech., DESRIST 2020, Kristiansand, Norway, Dec. 2--4, 2020, Proc. 15}.\hskip 1em plus 0.5em minus 0.4em\relax Springer, 2020, pp. 15--27.

\bibitem{deittouvron2021training}
H.~Touvron, M.~Cord, M.~Douze, F.~Massa, A.~Sablayrolles, and H.~J{\'e}gou, ``Training data-efficient image transformers \& distillation through attention,'' in \emph{Int'l Conf. Mach. Learning}.\hskip 1em plus 0.5em minus 0.4em\relax PMLR, 2021, pp. 10\,347--10\,357.

\bibitem{debertahe2020deberta}
P.~He, X.~Liu, J.~Gao, and W.~Chen, ``Deberta: Decoding-enhanced bert with disentangled attention,'' \emph{arXiv preprint arXiv:2006.03654}, 2020.

\bibitem{WAT1woodbury2004development}
M.~G. Woodbury, P.~E. Houghton, K.~E. Campbell, and D.~H. Keast, ``Development, validity, reliability, and responsiveness of a new leg ulcer measurement tool,'' \emph{Advances in Skin \& Wound Care}, vol.~17, no.~4, pp. 187--196, 2004.

\bibitem{10huang2019clinicalbert}
K.~Huang, J.~Altosaar, and R.~Ranganath, ``Clinicalbert: Modeling clinical notes and predicting hospital readmission,'' \emph{arXiv preprint arXiv:1904.05342}, 2019.

\bibitem{perez2018data}
F.~Perez, C.~Vasconcelos, S.~Avila, and E.~Valle, ``Data augmentation for skin lesion analysis,'' in \emph{Third International Workshop, ISIC 2018, Held in Conjunction with MICCAI 2018, Granada, Spain, September 16 and 20, 2018, Proceedings 5}.\hskip 1em plus 0.5em minus 0.4em\relax Springer, 2018, pp. 303--311.

\bibitem{openai2023gpt4}
\BIBentryALTinterwordspacing
OpenAI, ``Gpt-4 technical report,'' 2023, accessed: 2024-08-14. [Online]. Available: \url{https://openai.com/research/gpt-4}
\BIBentrySTDinterwordspacing

\bibitem{20thompson2013reliability}
N.~Thompson, L.~Gordey, H.~Bowles, N.~Parslow, and P.~Houghton, ``Reliability and validity of the revised photographic wound assessment tool on digital images taken of various types of chronic wounds,'' \emph{Advances in skin \& wound care}, vol.~26, no.~8, pp. 360--373, 2013.

\bibitem{deit_keras}
\BIBentryALTinterwordspacing
K.~Team, ``Data-efficient image transformers (deit) - keras example,'' 2023, accessed: 2024-10-27. [Online]. Available: \url{https://keras.io/examples/vision/deit/}
\BIBentrySTDinterwordspacing

\bibitem{zhang2017mixup}
H.~Zhang, ``mixup: Beyond empirical risk minimization,'' \emph{arXiv preprint arXiv:1710.09412}, 2017.

\bibitem{yun2019cutmix}
S.~Yun, D.~Han, S.~J. Oh, S.~Chun, J.~Choe, and Y.~Yoo, ``Cutmix: Regularization strategy to train strong classifiers with localizable features,'' in \emph{Proc. IEEE/CVF Int'l Conf. Comp. Vision}, 2019, pp. 6023--6032.

\bibitem{zhong2020random}
Z.~Zhong, L.~Zheng, G.~Kang, S.~Li, and Y.~Yang, ``Random erasing data augmentation,'' in \emph{Proc. AAAI Conf. Artificial intelligence}, vol.~34, no.~07, 2020, pp. 13\,001--13\,008.

\bibitem{qian2022limitations}
J.~Qian, H.~Wang, Z.~Li, S.~Li, and X.~Yan, ``Limitations of language models in arithmetic and symbolic induction,'' \emph{arXiv preprint arXiv:2208.05051}, 2022.

\bibitem{intermediateFusionyoo2019deep}
Y.~Yoo, L.~Y. Tang, D.~K. Li, L.~Metz, S.~Kolind, A.~L. Traboulsee, and R.~C. Tam, ``Deep learning of brain lesion patterns and user-defined clinical and mri features for predicting conversion to multiple sclerosis from clinically isolated syndrome,'' \emph{Comp. Methods in Biomechanics and Biomedical Engr: Imaging \& Vis.}, vol.~7, no.~3, pp. 250--259, 2019.

\bibitem{liu2023chronic}
Z.~Liu, E.~Agu, P.~Pedersen, C.~Lindsay, B.~Tulu, and D.~Strong, ``Chronic wound image augmentation and assessment using semi-supervised progressive multi-granularity efficientnet,'' \emph{IEEE Open Journal of Engineering in Medicine and Biology}, 2023.

\end{thebibliography}

%








\end{document}